\documentclass[fleqn,10pt]{wlscirep}
\usepackage[utf8]{inputenc}
\usepackage[T1]{fontenc}
\usepackage{graphicx}

\title{Application of Tensorized Neural Networks for Cloud Classification}
\author[1,*]{Alifu Xiafukaiti}
\author[2]{Devanshu Garg}
\author[1]{Aruto Hosaka}
\author[1]{Koichi Yanagisawa}
\author[2]{Yuichiro Minato}
\author[1]{Tsuyoshi Yoshida}
\affil[1]{Information Technology R \& D Center, Mitsubishi Electric Corporation, 5-1-1, Ofuna, Kamakura, Kanagawa, 247-8501 Japan}
\affil[2]{blueqat Inc., 2-24-12 Shibuya, Shibuya-ku, Tokyo, 150-6139 Japan}
\affil[*]{Xiafukaiti.Alifu@ct.MitsubishiElectric.co.jp}
\DeclareUnicodeCharacter{2212}{-}
\begin{abstract}
Convolutional neural networks (CNNs) have gained widespread usage across various fields such as weather forecasting, computer vision, autonomous driving, and medical image analysis due to its exceptional ability to extract spatial information, share parameters, and learn local features. However, the practical implementation and commercialization of CNNs in these domains are hindered by challenges related to model sizes, overfitting, and computational time. To address these limitations, our study proposes a groundbreaking approach that involves tensorizing the dense layers in the CNN to reduce model size and computational time. Additionally, we incorporate attention layers into the CNN and train it using Contrastive self-supervised learning to effectively classify cloud information, which is crucial for accurate weather forecasting. We elucidate the key characteristics of tensorized neural network (TNN), including the data compression rate, accuracy, and computational speed. The results indicate how TNN change their properties under the batch size setting.
\end{abstract}
\begin{document}
\flushbottom
\maketitle
%
%
\thispagestyle{empty}
\section*{Introduction}
Convolutional neural networks (CNNs) are a type of deep learning model that has been extensively utilized in computer vision tasks, particularly in image recognition and processing \cite{Yamashita,Tuli,ZhangZ}. The fundamental concept behind CNNs is to exploit the spatial structure of data, such as images, by employing convolutional layers to extract local features through filters and pooling layers to downsample and capture higher-level features. CNNs have achieved remarkable success in experimental evaluations due to their advantages in hierarchical feature learning, parameter sharing, and translation invariance \cite{Hasani,Koushik,Taye}. However, CNNs involve computations like convolutions and pooling, which can be computationally demanding, especially for large-scale models or datasets. Moreover, CNNs often require significant computational resources, especially when dealing with large datasets or complex architectures. Training and inference can be computationally expensive, which may limit their practical use in resource-constrained environments.

Recent advancements in weather forecasting, particularly in cloud classification using artificial intelligence, have brought numerous benefits, and addressed certain challenges \cite{Kurihana1,Kurihana2}. Clouds are hydrometeors composed of tiny particles of liquid water or ice suspended in the atmosphere. Meteorologists create weather forecasts and gain insights into meteorological phenomena by analyzing the shape, structure, altitude, and other characteristics of clouds \cite{Baker,Boucher,Zelinka}. Therefore, cloud classification plays a crucial role in meteorology. For instance, clouds at different altitudes are categorized into various types based on their formation, enabling meteorologists to accurately predict cloud movement and changes in meteorological phenomena. Additionally, cloud classification is valuable in studying climate change \cite{Goren,Zheng,Liu}. Clouds are significant factors in Earth's radiation budget and can influence climate change by enhancing atmospheric cooling or warming effects. Cloud classification is a complex task due to factors such as shape, altitude, size, color, density, and movement \cite{Chen,Tapakis}. In the past, various methods were employed to classify clouds from camera images, including decision trees \cite{Buch}, K-nearest neighbors classifiers \cite{Wang}, support vector machines \cite{Li}, linear discriminant analysis \cite{Oikonomou}, and others. These methods are relatively simple to implement, work well with a small number of input variables, and can be used for both classification and regression tasks. They are also effective in high-dimensional spaces and allow for the specification of different kernel functions. However, they have certain drawbacks, such as being memory-intensive, difficult to interpret, and sensitive to the choice of kernel function and parameter settings. Additionally, they assume that the data is normally distributed and may not perform well with non-linear data. In comparison, CNN methods have gained popularity in image recognition, object detection, and other domains \cite{Khanafer,Wen,Lin}. CNN models can handle large amounts of data and can be trained to recognize complex patterns within the data. Moreover, they have demonstrated high accuracy in classifying clouds in sky images \cite{LiuCC,Lv}. Nevertheless, there are some disadvantages to using deep learning for cloud classification. One major drawback is that CNN models require substantial amounts of labeled data and computational time for training, which can be time-consuming and expensive to acquire. Additionally, CNN models can be computationally demanding and necessitate powerful hardware for training and execution. Tensorized neural networks (TNNs) are powerful tools for efficiently handling multidimensional arrays called tensors and expressing mathematical operations concisely. Their advantages, such as overcoming the curse of dimensionality, reducing parameters, exploiting data locality, being resilient to noise, and allowing for physical interpretation, make tensorized networks valuable in machine learning, quantum information processing, and statistical physics \cite{Sengupta,Rieser,Ran}. 
 
In this study, our focus is on reducing the size and computational time by tensorizing the dense layers in the CNN. This addresses the limitations of practical implementation and commercialization of CNNs in cloud classification. Additionally, we incorporated attention layers into the CNN and trained it using contrastive self-supervised learning (CSSL) to effectively classify cloud information, which is crucial for accurate weather forecasting. The experiments are conducted using a type of CNN model known as densely connected convolutional networks (DenseNet) \cite{ref-Densenet}. DenseNet is chosen due to its distinctive characteristic of densely connecting layers, where each layer is connected to all other layers through feed-forward connections. This unique architecture makes DenseNet highly proficient in tasks related to image recognition. The tensorized layer models used in DenseNet achieved data compression rates of 95.4\% and a maximum increase in computational speed of 22.4\%. The result of experiments on two different Graphics processing units (GPUs) reveal that the accuracies remain consistent when using batch sizes smaller than the number of streaming multiprocessors (SMs) on the GPU, for both tensorized and general models. However, for batch sizes larger than the number of SMs on the GPU, the tensorized model exhibits improved accuracies.
\section{Methodology}
\subsection*{Cirrus Cumulus Stratus Nimbus Database}
\begin{figure}[h]
	\centering
	\includegraphics[width=0.55\linewidth]{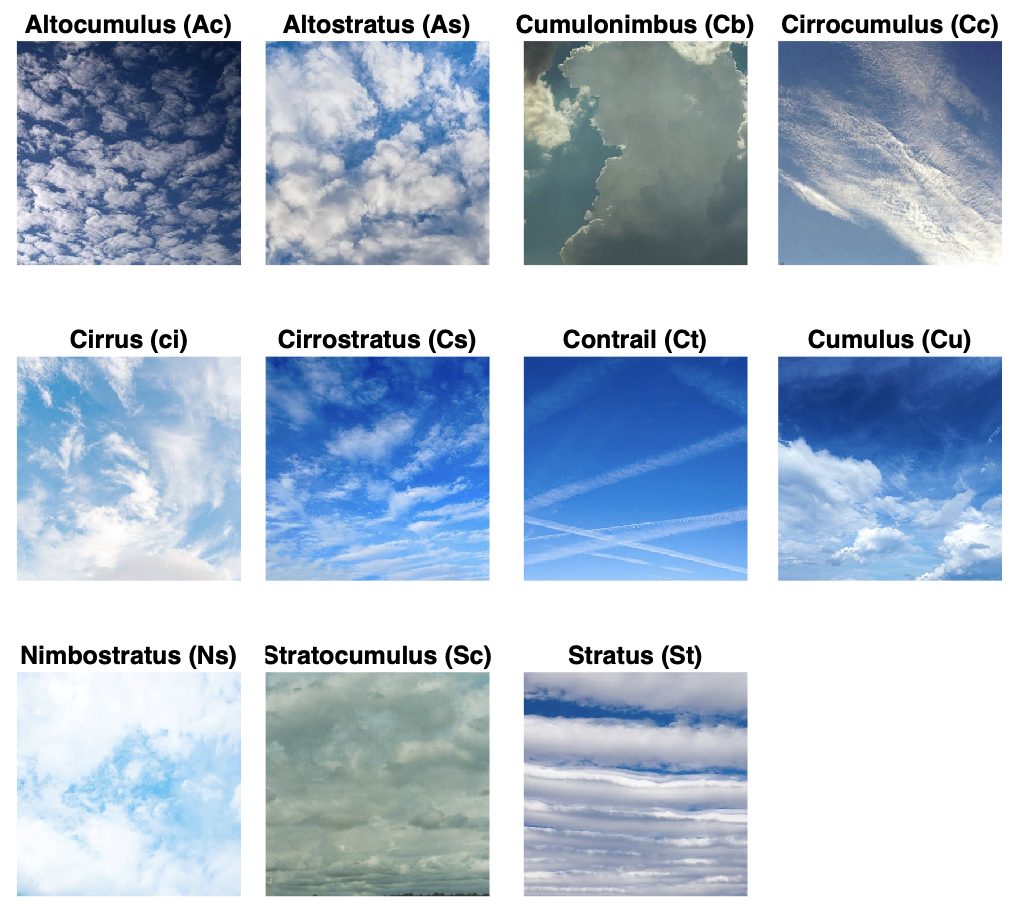}
	\caption{Representative sample images of 11-cloud categories from CSSN database. 
		\label{Fig1}}
\end{figure}

The Cirrus cumulus stratus nimbus (CCSN) database was a ground-based cloud database developed by Zhang’s team at Nanjing university of information science and technology \cite{ref-Dataset1}. Images in the CCSN database was divided into 11 different cloud categories based on the world meteorological organization’s genera-based classification recommendation \cite{ref-Dataset2}, including Altocumulus (Ac), Altostratus (As), Cumulonimbus (Cb), Cirrocumulus (Cc), Cirrus (Ci), Cirrostratus (Cs), Contrail (Ct), Cumulus (Cu), Nimbostratus (Ns), Stratocumulus (Sc), and Stratus (St). One representation from each class is shown in the Fig. \ref{Fig1}, and the sample sizes of each class are delineates at Table\ref{tab1}. The CCSN database provides labeled images with the corresponding cloud types, facilitating the development and evaluation of automated cloud classification algorithms. This extensive database encompasses a wealth of image data collected from satellites and ground sensors. Consequently, it enables the utilization of statistical analysis and machine learning algorithms for training purposes, thereby facilitating in-depth research on the distinct characteristics and behaviors exhibited by various cloud formations.All images are fixed resolution 256 × 256 pixels with the JPEG format. The CCSN database contains 2543 cloud images, which belong to a light-scale database. 
\begin{table}[h]
	\centering
	\begin{tabular}{|l|l|l|}
		\hline
		\textbf{Categories}	& \textbf{Abbreviation} & \textbf{Sample Size} \\
		\hline
		Altocumulus	& Ac & 221 \\
		\hline
		Altostratus	& As & 188 \\
		\hline
		Cumulonimbus & Cb & 242 \\
		\hline
		Cirrocumulus & Cc & 268 \\
		\hline
		Cirrus	& Ci & 139 \\
		\hline
		Cirrostratus & Cs & 287 \\
		\hline
		Contrail & Ct & 200 \\
		\hline
		Cumulus	& Cu & 182 \\
		\hline
		Nimbostratus & Ns & 274 \\
		\hline
		Stratocumulus & Sc & 340 \\
		\hline
		Stratus	& St & 202 \\
		\hline
		Total & - & 2543 \\
		\hline
	\end{tabular}
	\caption{\label{tab1}Categories, abbreviations, and sample sizes of CCSN Database.}
\end{table}

\subsection*{Contrastive Self-Supervised Learning}
\begin{figure}[]
 	\centering
 	\includegraphics[width=0.8\linewidth]{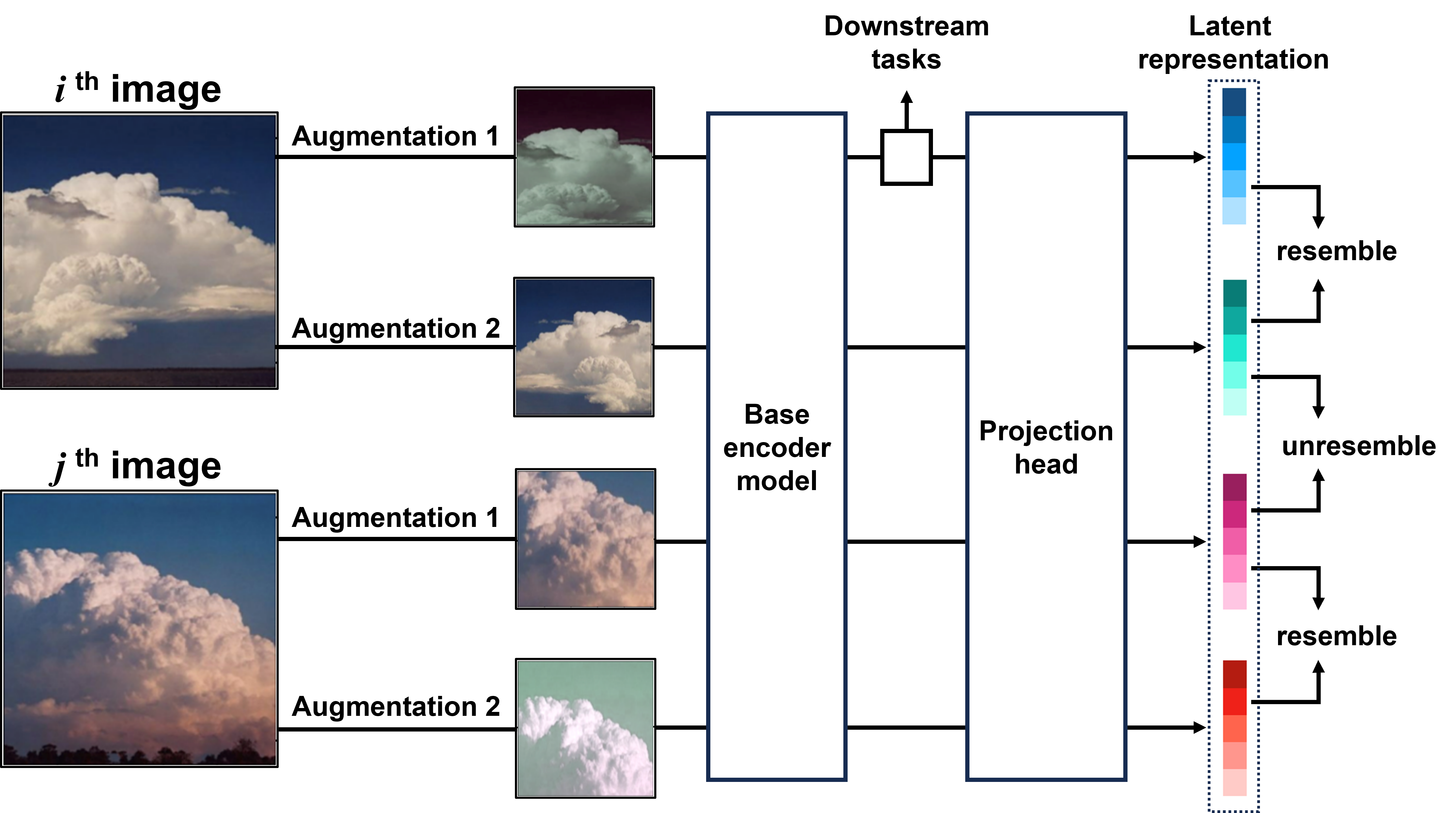}
 	\caption{The pipeline for CSSL involves the following steps: Each image generates two augmentations, which are then fed into the base model for feature representation. These representations are subsequently passed to the projection head, which produces the latent representation of the images. The objective is to minimize the distance between these representations for the same image. Once training is complete, the projection head is removed, and the output from the base model is utilized for downstream tasks. \cite{ref-simclr}
 		\label{Fig2}}
 \end{figure}

To maximize the utilization of a small sample size, we employ contrastive learning techniques\cite{ref-CSSL1}. By leveraging unlabeled data, the model learns to identify differences in data features without relying on labels. This approach has been proven effective in various tasks such as image recognition, natural language processing, and speech processing \cite{ref-CSSL2, ref-CSSL3, ref-CSSL4}. Our network consists of two models: a base encoder for generating features and a projection head for creating latent space representations. During training, two random augmentations are applied to each image in a batch. The model is then optimized by ensuring that the latent space representations of the augmentations are similar for the same image, while being dissimilar for different images, as shown in Fig. \ref{Fig2}. The best combination of augmentations were found to be when colour jitter, crop, and resizing were applied at random \cite{ref-simclr} from an extensive array of different augmentations. After training, a portion of the projection head is removed, and a new classifier head is attached to this modified model for downstream tasks. The tensorization process takes place in the first dense layer of the projection head, which contains the majority of parameters.
\subsection*{Tensorization of Dense Layer}
While wide networks are necessary for problems involving large datasets or sensitive features, it is observed that many parameter updates during training are redundant \cite{ref-NNred}. TNN have been successfully applied in various applications, including image classification \cite{ref-CV1, ref-CV2}, temporal prediction tasks \cite{ref-TP1}, and large language modeling \cite{ref-LM1, ref-LM2}. Different forms of decomposition, such as CANDECOMP/PARAFAC (CP) \cite{ref-CPD1, ref-CPD2}, tucker decomposition \cite{ref-TD1}, and tensor train or matrix product states-based methods \cite{ref-TT1, ref-TT2}, have been extensively tested for almost every component of state-of-the-art models.

\begin{figure}[]
	\centering
	\includegraphics[width=0.5\linewidth]{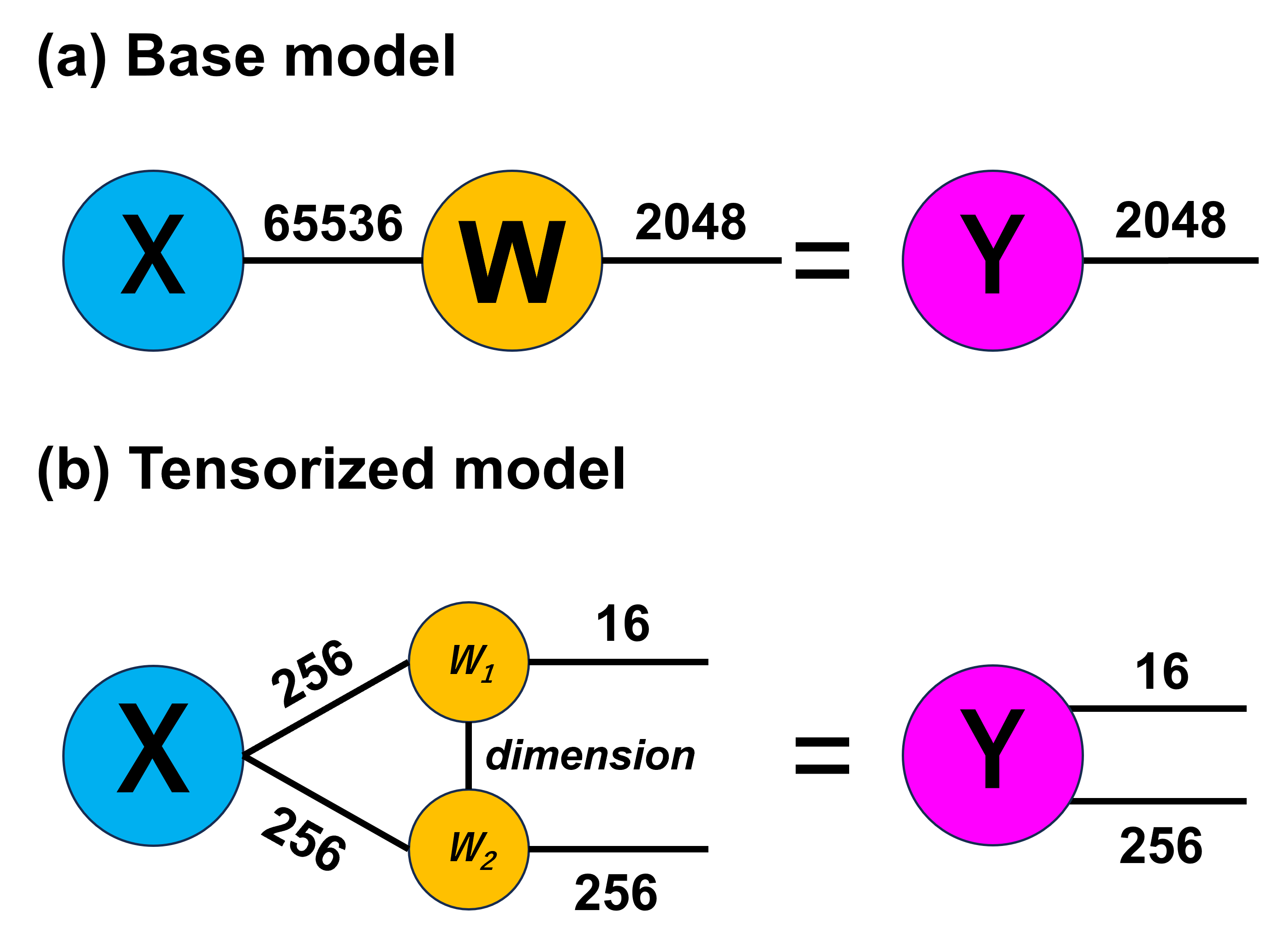}
	\caption{The tensor decomposition scheme is as follows: (a) The input X is obtained from the base model after flattening. W represents the weight matrix of the first dense layer. The shared index denoted by a line indicates contraction, resulting in the output Y. This output serves as the input for the subsequent dense layer. (b) The input tensor X is reshaped into a rank-2 tensor. The weight matrix is now represented by two smaller rank-3 weight matrices. This approach maintains the same output dimensionality while reducing the number of parameters required to represent the original weight tensor.
	\label{Fig3}}
\end{figure}

In this study, a significant portion of the parameters in the models were concentrated in the dense layers. To address this, we employed a simple tensor rain decomposition technique for the weight matrix of the first dense layer. This involved breaking down the large rank-2 weight matrix into two smaller rank-3 weight matrices, connected through a shared index, as illustrated in Fig.\ref{Fig3}. The size of this index is referred to as the bond dimension, which influences the representation power. Despite having the same input and output dimensions, the tensorized dense layer utilized significantly fewer parameters.
Table\ref{tab2} shows the reduction in parameters achieved using TNN at various bond dimensions of the weight tensor. Notably, at a bond dimension of 16, the rate of parameter reduction is remarkably high, reaching 95.4\%. However, as the bond dimension increases from 32 to 256, this reduction rate decreases. Nevertheless, TNN consistently achieves a reduction of over 90\% in parameters across all tested bond dimensions of the weight tensor. These findings underscore the effectiveness of TNN in significantly reducing the number of parameters required for the model. Even at higher bond dimensions, TNN still achieves substantial parameter reduction, indicating its potential for efficient and resource-saving computations.
\begin{table}[]
	\centering
	\begin{tabular}{|l|l|l|l|l|l|}
		\hline
		\textbf{Bond dimension of weight tensors}	& 16 & 32 & 64 & 128 & 256 \\
		\hline
		\textbf{Parameter reduction}	& 95.4\% & 95.0\% & 94.2\% & 92.6\% & 89.4\% \\
		\hline
	\end{tabular}
	\caption{\label{tab2}Reduction in parameters achieved using TNN at various bond dimensions of the weight tensor. }
\end{table}
\subsection*{Implementation and Hyper-parameters}
All models are implemented using TensorFlow. The DenseNet used as encoder model, obtained directly from the keras repository. The decomposed tensor layer is also implemented in TensorFlow using the keras layer subclass. Training is performed on a cluster consisting of eight A100 nodes and four V100 nodes. The A100 is a high-performance GPU designed for artificial intelligence (AI) and scientific computing workloads. This configuration allows for significant parallel processing power and is commonly used in applications such as deep learning, data analytics, and high-performance computing. The V100 is a powerful GPU designed for AI, machine learning, and high-performance computing tasks. This configuration provides substantial parallel processing capabilities and is commonly used in applications such as deep learning training and inference, scientific simulations, and data analytics. During the initial phase of contrastive learning, the entire dataset is utilized. In the subsequent label learning phase, an 80:20 split is employed for training and validation. The ADAM optimizer is used, along with an exponential learning rate scheduler with 80,000 decay steps and an initial learning rate of 0.02. The structure of the projection head and classifier head remained consistent across all models, with sequential layers containing [4096, 1024, 512] neurons for the projection head and [4096, 11] neurons for the classifier head.
\subsubsection*{Algorithm Flow}
The whole process is summarised through the following steps:
\begin{itemize}
	\item A composite model of encoder model trained on ImageNet and projection head with random weights is initialised.
	\item For the contrastive learning phase the encoder model is kept frozen for 50 epochs.
	\item Now the encoder model is unfreezed and the whole model is trained fo the rest of the epochs.
	\item After the contrastive learning phase is completed, the last two layers of the projection head are removed. The snipped model is now used for down-stream tasks. A classifier head is attached for label learning.
	\item Now this model is trained on the split dataset (training dataset) using supervised learning.
\end{itemize}
This process is done for DenseNet and it’s tensorised versions. Although the TNNs have significantly less parameters, the method of their decomposition affects the increase in computational speed (Henceforth, it shall be referred to as speedup) observed (e.g. speedup of 5\%, 4\% and 1\% are obtained from output dimension of (16, 256), (32, 128), (64, 64) for weight heads, respectively). Here, the speedup is defined as the rate of reduction in computational time. Notably, there is a significant increase in speedup when using output dimensions of (16, 256) for weight heads. However, the speedup decreases after using output dimensions of (64, 64) for weight heads. This highlights the importance of carefully selecting the dimensions of the heads and inputs. The findings emphasize that the choice of output dimensions for weight heads plays a critical role in optimizing the model's performance. It is crucial to consider the relationship between the dimensions of the heads and the inputs to ensure efficient and effective computation. By selecting appropriate dimensions, we can maximize both the speedup and overall performance of the model. 
\subsubsection*{Loss Function}
The loss function used in this paper has no modification from the inspired work \cite{Chen}. For a batch size of $N$ images, defining the contrastive distance between a pair of augmented images produces 2$N$ data points. For a positive pair, the other $2(N − 1)$ augmented examples are considered as negative examples. Let $sim(u, v) = u^T v/\parallel u\parallel \parallel v\parallel$ denote the dot product between $l_2$ normalized $u$ and $v$ (i.e.cosine similarity). Then the loss function for a positive pair of examples ($i$, $j$) is defined as
\begin{equation}
	l_{i,j} = − $log$ \frac{exp(sim(z_i, z_j )/\tau)}
	{\sum_{k=1}^{2N} 1_{[k\neq i]}exp(sim(z_i, z_j )/\tau)},
\end{equation}
where $1_{[k\neq i]}$ $\in {0, 1}$ is an indicator function evaluating to 1 iff $k\neq i$ and $\tau$ denotes a temperature parameter. The final loss is computed across all positive pairs, both ($i$, $j$) and ($j$, $i$), in a mini-batch. This loss has been used in previous works \cite{ref-simclr, ref-Loss1, ref-Loss2}
\section*{Results and Discussion}
\begin{figure}[]
	\centering
	\includegraphics[width=\linewidth]{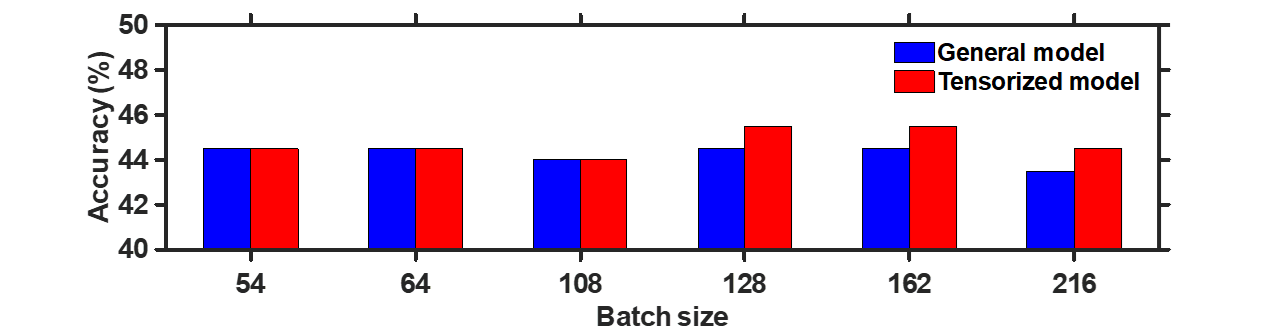}
	\caption{The Top-1 accuracy observed using a node of 8 × A100. 
		\label{Fig4}}
\end{figure}

\begin{figure}[]
	\centering
	\includegraphics[width=\linewidth]{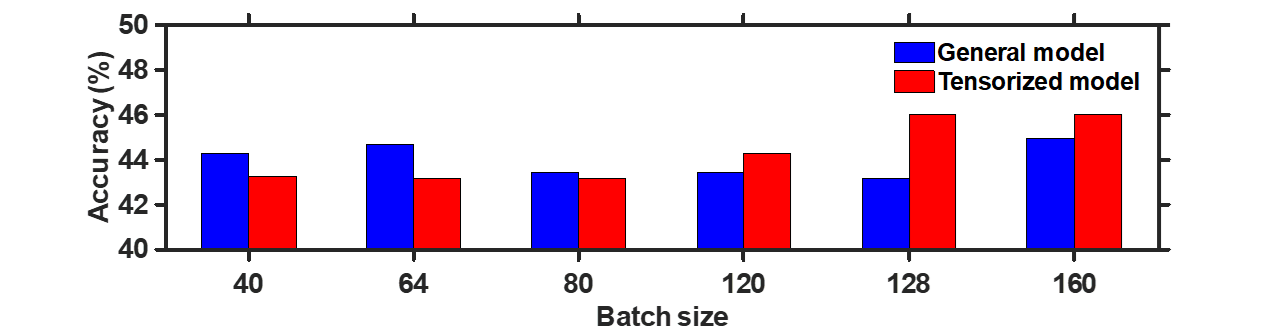}
	\caption{The Top-1 accuracy observed using a node of 4 × V100.
		\label{Fig5}}
\end{figure}

The cloud classification training involves the utilization of DenseNet with both a general model and a tensorized model in two distinct environments. The computational environment consists of a configuration comprising 8 × A100 and 4 × V100 nodes, which refers to a computing system or server setup equipped with eight NVIDIA A100 and four NVIDIA V100 GPUs. The Top-1 accuracy of these models across different batch sizes for each environment is depicted in the bar graphs shown in Figs. \ref{Fig4} and \ref{Fig5}. In general, a GPU is composed of multiple streaming multiprocessors (SMs) that enable parallel processing. Each SM is responsible for concurrently executing multiple thread blocks. To consider their respective SM capabilities (A100 having 108 SMs and V100 having 80 SMs) and reduce learning instability, the batch size starts from half of the GPU's SMs and stops at two times the GPU's SMs. When training on 8 × A100, both models achieve similar accuracies when the batch size is less than 108. The tensorized model outperforms the general model in terms of accuracy when the batch size is greater than 108. When training on 4 × V100, the general model achieves higher accuracy when the batch size is less than 80. The tensorized model surpasses the general model in terms of accuracy when the batch size is greater than 80. It appears that, in the case of the tensorized model, higher accuracy is achieved when the batch sizes exceed the number of SMs on the GPUs. With a smaller batch size, the model can observe a greater variety of data points and construct a more generalized model. However, training with a batch size that exceeds the number of GPU SMs can lead to increased training instability and inaccurate gradient estimation. It can also impede the convergence of the training process, resulting in convergence to a local optimum and a subsequent decrease in accuracy after training. However, by utilizing techniques such as TNN, data reduction can be achieved, enabling the mitigation of training instability even with larger batch sizes. This suggests that utilizing larger batch sizes allows for better utilization of the available computational resources on the GPUs, leading to improved accuracy in the TNN. It is quite remarkable, especially considering the significant reduction in model size, amounting to almost 95.4\%.

\begin{figure}[]
	\centering
	\includegraphics[width=\linewidth]{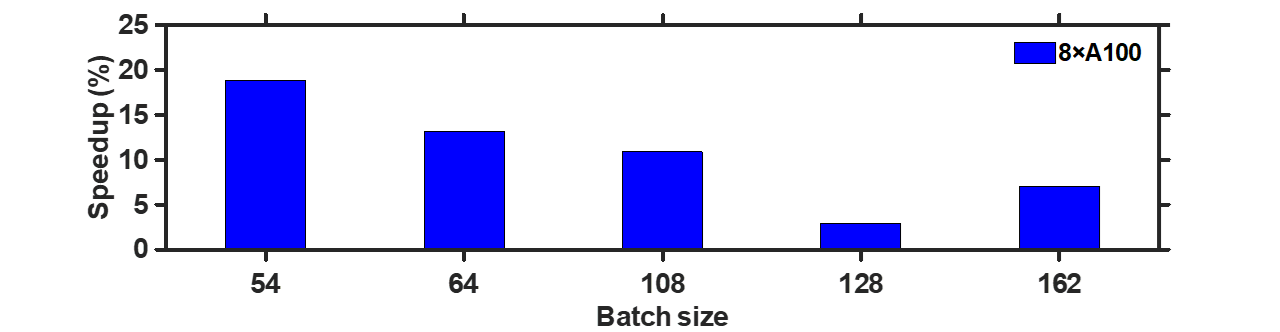}
	\caption{The speedup observed for the TNN using a node consisting of 8 × A100 is significant. The reported speedup is observed when compared to the general model using the same batch size.
		\label{Fig6}}
\end{figure}

\begin{figure}[]
	\centering
	\includegraphics[width=\linewidth]{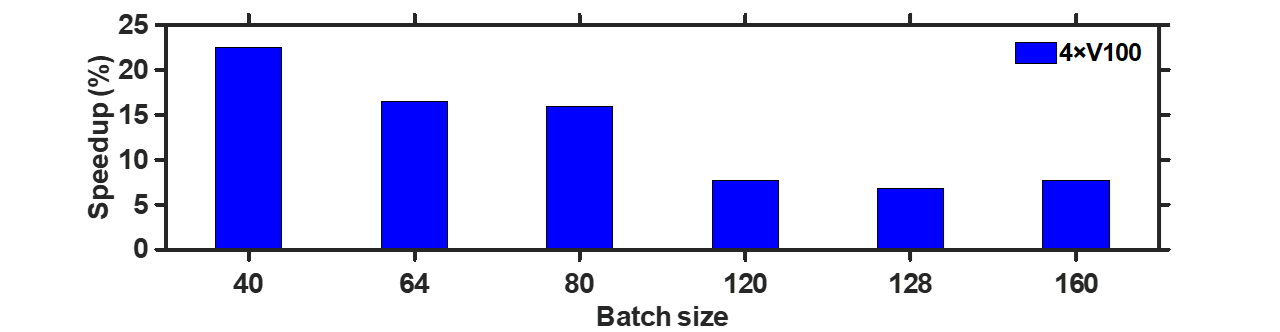}
	\caption{The speedup observed for the TNN using a node consisting of 4 × V100 is significant. The reported speedup is observed when compared to the general model using the same batch size .\label{Fig7}}
\end{figure}

To demonstrate the impact of TNN on speedup across different batch sizes, Figs \ref{Fig6} and \ref{Fig7} depict the speedup achieved by TNN training compared to CNN using 8 × A100 and 4 × V100 nodes, respectively. The tensorized layer models are tested with a batch size of 54 for 8 × A100 and 40 for 4 × V100, resulting in a maximum speedup of 22.4\% and 18.9\%, respectively. In both computational environments, when using a batch size lower than the number of SMs on the GPUs, TNN exhibited a computational speed increase of over 10\% per training iteration compared to traditional CNN. This is because, within this batch size range, the maximum number of GPU iterations for a single batch remains the same. However, the speedup diminishes when the batch size exceeds the number of SMs on the GPUs. Each SM on a GPU can execute multiple thread blocks concurrently. However, with a large batch size, the allocation of thread blocks to each SM decreases, limiting the GPU's parallel processing capability and potentially prolonging computation time. Nevertheless, TNN can still enhance computation speed even with larger batch sizes. Overall, employing a smaller batch size in TNN enables more efficient computation, resulting in faster training times. It is worth noting that there is a slight trade-off in accuracy compared to traditional CNN. However, the benefits of reduced computational time may outweigh the slight decrease in accuracy for certain applications or scenarios.
\section*{Conclusions}
To achieve real-time cloud classification, our study proposes a groundbreaking approach that involves tensorizing the dense layers in the CNN to reduce model size and computational time. By investigating the key characteristics of TNN, including data compression rate, accuracy, and computational speed, we are able to achieve significant reductions in both model size and computational time. The tensorized layer models used in DenseNet achieved impressive data compression rates of 95.4\% and a maximum increase in computational speed of 22.4\%. Experimental results on two different GPUs consistently demonstrated high accuracies when using batch sizes smaller than the number of SMs on the GPU, for both tensorized and general models. Furthermore, computational speeds are notably higher within this range of batch sizes. This mean employing a smaller batch size in TNN enables more efficient computation, resulting in faster training times. In summary, this paper highlights the revolutionary potential of TNNs in real-time cloud classification and provides a detailed and comprehensive view of the future direction of meteorological research. 
\section*{Acknowledgments}
The CCSN database used in our paper can be found at https://github.com/upuil/CCSN-Database. We also thank the author who provided the CCSN database.
\bibliography{sample}
\noindent 
\end{document}